\documentclass[conference]{IEEEtran}
\IEEEoverridecommandlockouts
\usepackage{cite}
\usepackage{amsmath,amssymb,amsfonts}
\usepackage{algorithmic}
\usepackage{graphicx}
\usepackage{textcomp}
\usepackage{xcolor}
\usepackage{graphicx}
\usepackage{caption}  
\usepackage{comment}
\usepackage[colorlinks=true, linkcolor=blue, citecolor=blue, urlcolor=blue]{hyperref}
\def\BibTeX{{\rm B\kern-.05em{\sc i\kern-.025em b}\kern-.08em
    T\kern-.1667em\lower.7ex\hbox{E}\kern-.125emX}}
\begin{document}

\title{VisNet: Efficient Person Re-Identification via $\alpha$-Divergence Loss, Feature Fusion and Dynamic Multi-Task Learning}

\author{\IEEEauthorblockN{Anns Ijaz}
\IEEEauthorblockA{\textit{Department of Artificial Intelligence} \\
\textit{University of Management and Technology}\\
Lahore, Pakistan \\
annsijaz@outlook.com}
\and
\IEEEauthorblockN{Dr. Muhammad Azeem Javed}
\IEEEauthorblockA{\textit{Department of Artificial Intelligence} \\
\textit{University of Management and Technology}\\
Lahore, Pakistan \\
azeem.javed@umt.edu.pk}
}

\maketitle

\begin{abstract}
Person re-identification (ReID) is an extremely important area in both surveillance and mobile applications, requiring strong accuracy with minimal computational cost. State-of-the-art methods give good accuracy but with high computational budgets. To remedy this, this paper proposes VisNet, a computationally efficient and effective re-identification model suitable for real-world scenarios. It is the culmination of conceptual contributions, including feature fusion at multiple scales with automatic attention on each, semantic clustering with anatomical body partitioning, a dynamic weight averaging technique to balance classification semantic regularization, and the use of loss function FIDI for improved metric learning tasks. The multiple scales fuse ResNet50's stages 1 through 4 without the use of parallel paths, with semantic clustering introducing spatial constraints through the use of rule-based pseudo-labeling. VisNet achieves 87.05\% Rank-1 and 77.65\% mAP on the Market-1501 dataset, having 32.41M parameters and 4.601 GFLOPs, hence, proposing a practical approach for real-time deployment in surveillance and mobile applications where computational resources are limited.
\end{abstract}

\section{Introduction}
The advent of intelligent video analytics has enabled the development of large-scale surveillance networks for applications like retail analytics, forensic investigation, and public safety. At the core of these systems is Person Re-identification (re-ID), the task of matching an individual's appearance across non-overlapping camera views. While crucial for scalable analytics, achieving robustness remains difficult due to significant appearance variations across disparate views. The visual signature of an identity is frequently compromised by dramatic changes in camera angles and body poses (viewpoint variation), or by crowded scenes that obscure discriminative cues (occlusion). Furthermore, photometric inconsistencies from lighting changes, scale variations due to differing camera distances, and temporal gaps significantly alter appearance. Traditional hand-crafted features, such as color histograms and edge-based descriptors \cite{DeMarsicoEtAl2014}, fail to capture the semantic richness required to overcome such variations, necessitating more robust feature learning approaches.

\begin{figure}[h]
    \centering
    \includegraphics[width=1.0\linewidth]{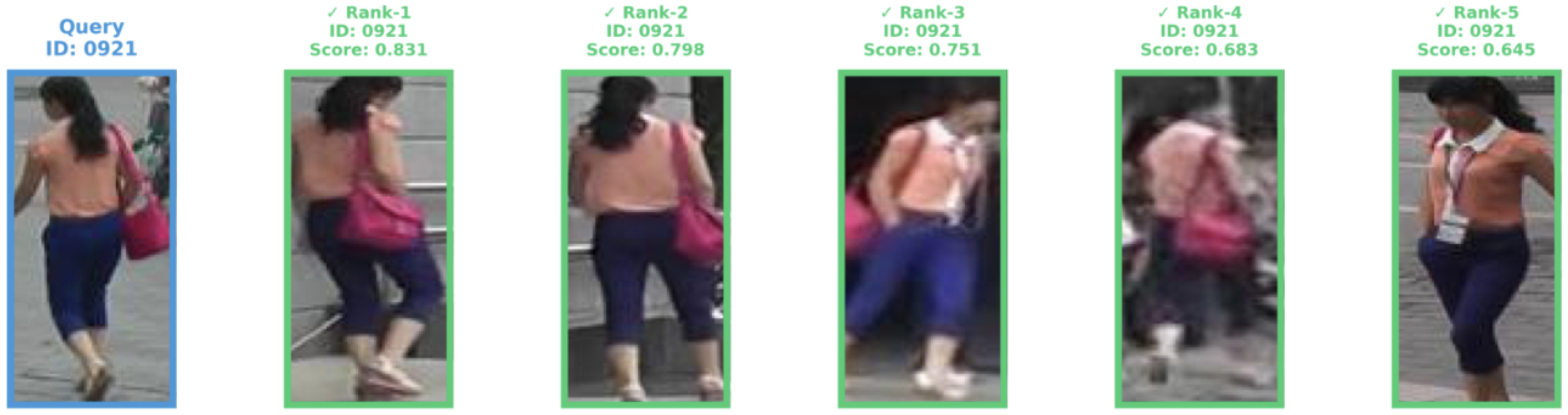}
    \caption{Market-1501 Query: The proposed model successfully re-identifies the person (ID:0921) across multiple viewpoints in the top-5 ranked results among 19,733 images.}
    \label{fig:single_col}
\end{figure}

With the advent of deep learning, spatial partition strategies and CNN-based methods offered a radical improvement. Methods like PCB and AANet captured local semantic structure by partitioning feature maps or introducing adaptive attention, achieving 76--92\% Rank-1 accuracy \cite{sun2018beyond} \cite{Tay2019}. However, these approaches remained inherently limited by their reliance on predefined spatial regions. Consequently, recent works have moved their focus towards Transformer architectures \cite{vaswani2017attention} and vision-language models to capture global semantic context. Solutions such as TransReID \cite{zhang2025efficient} and CLIP-ReID \cite{Li2023} report higher accuracy (greater than 88\% Rank-1) but are computationally prohibitive. For instance, TransReID introduces approximately 17.8 GFLOPs and 86M parameters, making its deployment on edge devices infeasible. This explicitly establishes an efficiency-accuracy trade-off that lightweight CNN-based methods remain computationally efficient but lack the semantic reasoning of Transformers, while Transformer-based models advance accuracy at a computational cost that prohibits deployment in resource-constrained environments.

To address this trade-off, a method is required that retains the spatial efficiency of CNNs while mimicking the semantic capture of Transformers. While multi-scale feature learning is well-known in computer vision, where spatial pyramid pooling and feature pyramid networks \cite{lin2017feature} improve object detection, it remains underutilized in re-ID. Furthermore, while attention mechanisms like SE-Net \cite{hu2018squeeze} and CBAM \cite{woo2018cbam} operate on channel or spatial dimensions, the scale dimension remains an under-explored design space about determining which multi-scale feature representation is most informative for a given image.

In this work, we propose VisNet, a systematically designed person re-ID model that achieves competitive accuracy while maintaining practical computational efficiency. Instead of utilizing computationally expensive Transformer architectures, VisNet leverages a strategic combination of proven CNN-based techniques. It extracts features from multiple ResNet50  residual blocks \cite{he2016deep} and projects them to a unified dimension. These are combined via a novel custom scale attention mechanism, a lightweight module that learns adaptive per-scale weighting to capture semantic information at multiple levels of abstraction. To further regularize semantic learning without expensive teacher-student frameworks like SOLIDER \cite{chen2023solider}, we introduce spatial semantic clustering with rule-based pseudo-labels (classifying upper body, lower body, and shoes). Finally, to ensure robust metric learning, we employ FIDI loss, Semantic Loss and Cross Entropy Loss with dynamic weight averaging \cite{liu2019multitask}, which balances the convergence rates of identity classification, metric learning, and semantic regularization tasks.

The proposed scheme is validated on the Market-1501 benchmark \cite{zheng2015scalable}, achieving 87.05\% Rank-1 accuracy and 77.65\% mAP with only 4.601G FLOPs and 31.08M parameters (0.36$\times$ the size of TransReID). The main contributions of this paper are listed as follows:

\begin{itemize}
    \item A lightweight, yet accurate CNN-based person re-ID model utilizing customized scale attention mechanism for learning adaptive per-scale weighting towards multi-scale feature fusion
    \item Demonstration of the approach that rule-based spatial pseudo-labels effectively regularize semantic learning without expensive teacher-student frameworks and serve as a competitive, simpler alternative to recent complex methods.
    \item Extensive empirical validation quantifying the contribution of each component, efficiency-accuracy trade-off analysis, and qualitative analysis of results.
    \item A semantic-aware augmentation framework that enforces background invariance by explicitly decoupling foreground identity from environmental clutter.
\end{itemize}

The paper is organized as follows: Section II covers the proposed VisNet architecture. Section III describes the evaluation performance, while Conclusions and Future Work are provided in Section IV and V, respectively.

\section{Proposed Method}
\label{sec:method}
\begin{figure}[h]
    \centering
    \includegraphics[width=1.0\linewidth]{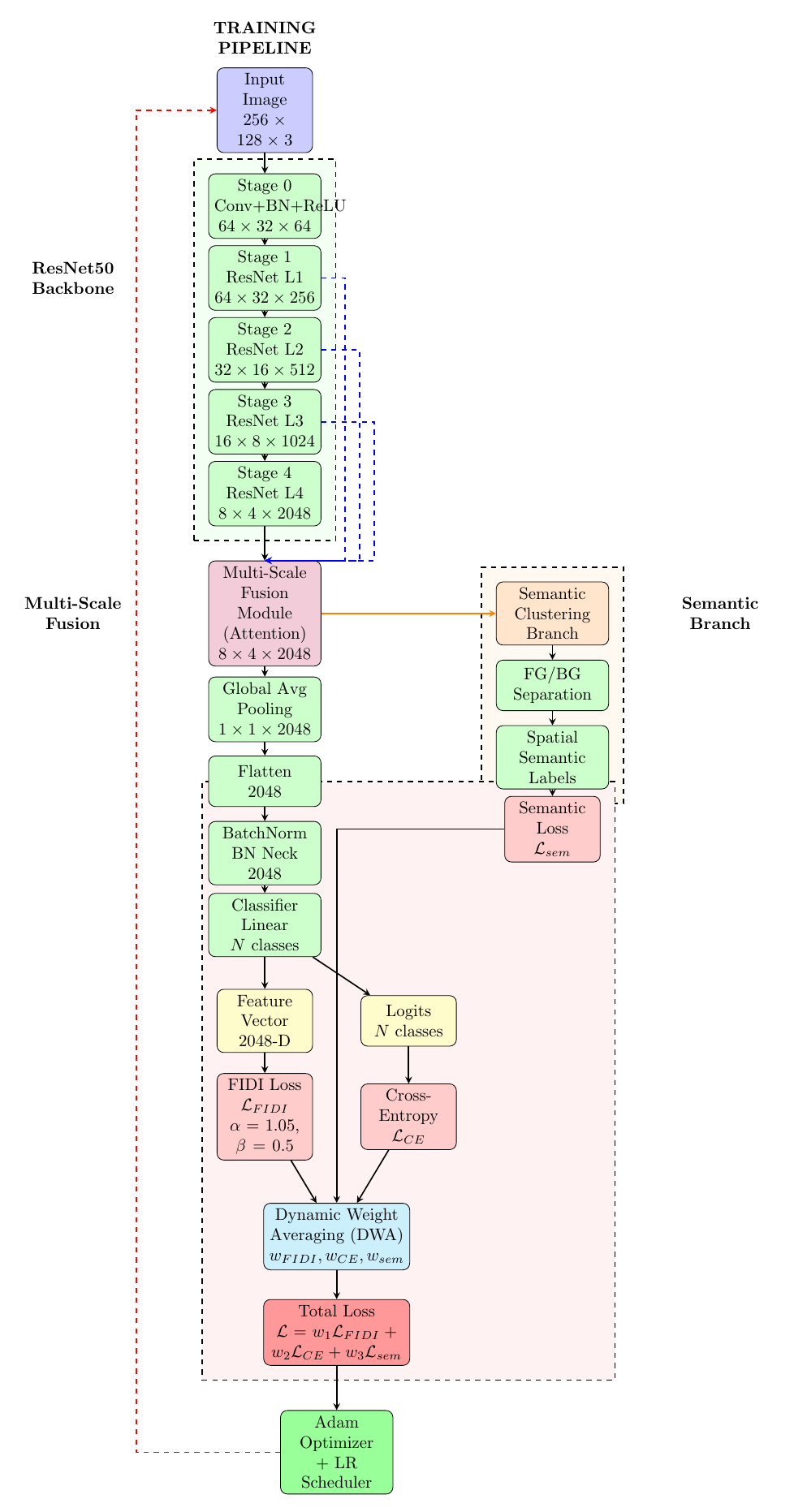}
    \caption{Training pipeline of the proposed method.}
    \label{fig:single_col}
\end{figure}

VisNet integrates a ResNet50 backbone, multi-scale feature fusion with learned attention, spatial semantic clustering, and a dynamic multi-task loss formulation. It processes the input $256 \times 128$ image through five stages of a ResNet50 backbone. Multi-scale features are extracted from stages 1 to 4 in different semantic levels and spatial resolutions. These are fused by a learned attention-weighted combination. These fused feature maps are then fed into two parallel heads, namely, an identity classification head for person re-identification and a semantic clustering head for spatial regularization. At inference time, only the identity classification head contributes toward the ultimate feature embedding. In training, both of them contribute toward the total loss through the framework of multi-task learning with dynamic weight averaging.

\begin{figure*}[t]
    \centering
    \includegraphics[width=\textwidth]{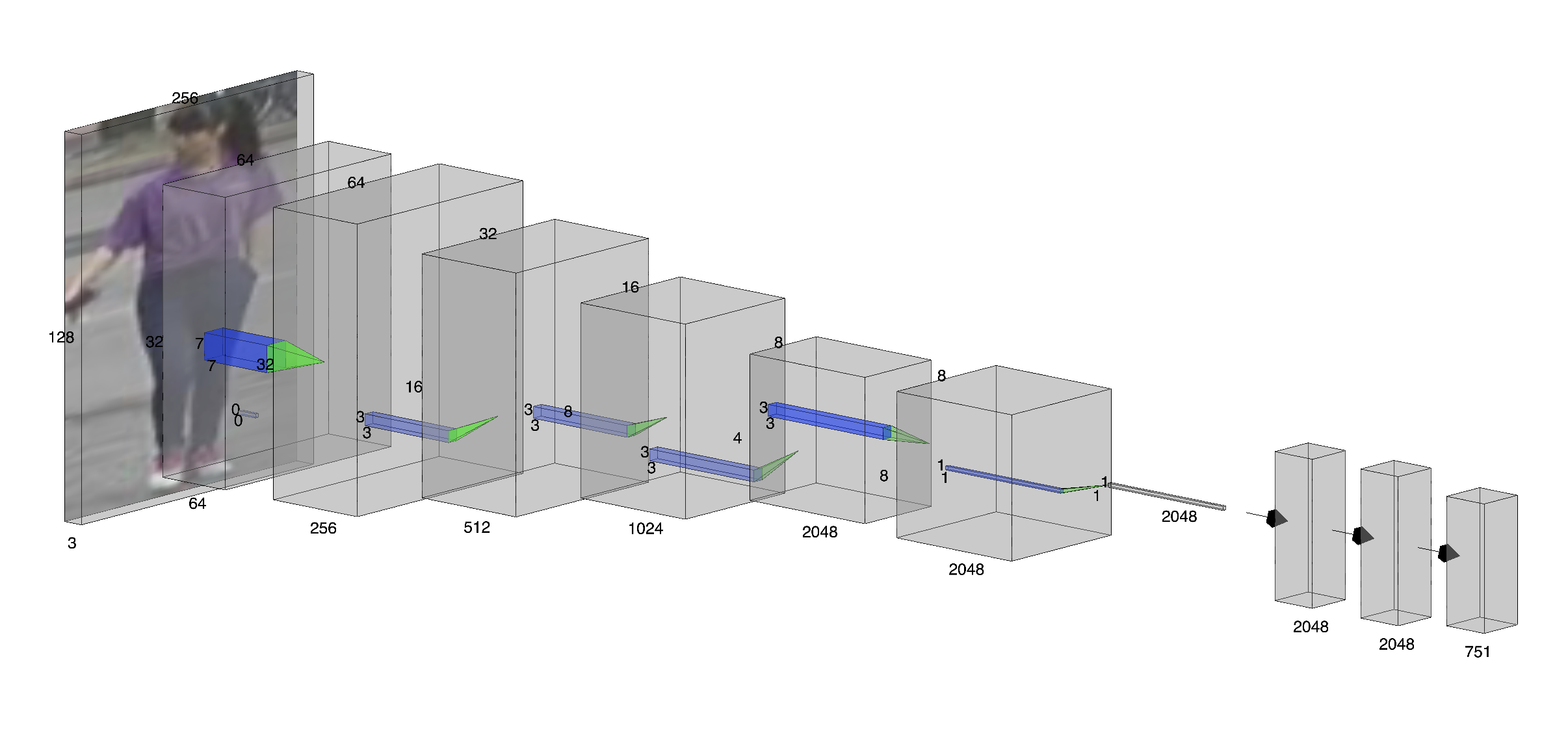}
    \caption{Proposed VisNet's Architecture}
    \label{fig:two_col}
\end{figure*}

\subsection{Backbone Architecture and Feature Extraction}

VisNet takes ResNet50 as backbone. The network is divided into five stages:

In detail, it is composed of: Stage 0: An initial stem of a $7 \times 7$ convolution with stride 2 is followed by batch normalization and ReLU\cite{ioffe2015batch} activation, then a subsequent max-pooling. This stage outputs 64 feature channels at a stride of 4.

\begin{itemize}
    \item \textbf{Stage 1--4:} Residual blocks corresponding to ResNet50's layer1 through layer4, which progressively extract features at increasing semantic abstraction levels.
\end{itemize}

The output feature dimensions for each stage are:
\begin{itemize}
    \item Stage 1 (layer1): 256 channels, stride 4
    \item Stage 2 (layer2): 512 channels, stride 8
    \item Stage 3 (layer3): 1024 channels, stride 16
    \item Stage 4 (layer4): 2048 channels, stride 32
\end{itemize}

Features from all four stages are extracted for multi-scale fusion, balancing computational efficiency with the need to capture both fine-grained and semantic-level information effectively.

\subsection{Multiscale Feature Fusion with Learned Scale Attention}

A central contribution of VisNet is the systematic fusion of multiscale features through learned per-scale attention weights; this is different from a simple concatenation or addition, where the relative informativeness of each ResNet stage for a given input image is determined.

\subsubsection{Projection to Unified Dimension}

Stages 1 through 4 have different channel dimensions: 256, 512, 1024, 2048. All features are projected to a common dimension of 2048 channels via $1 \times 1$ convolutions followed by batch normalization and ReLU activation:

\begin{equation}
F_i' = \text{ReLU}(\text{BN}(\text{Conv}_{1\times1}(F_i)))
\end{equation}

where $F_i$ is the feature map from stage $i$, and $F_i'$ its projected representation. Label smoothing with $\epsilon = 0.1$ is applied during training for improved generalization.

\subsubsection{Spatial Alignment}

These features are then bilinearly upsampled to the spatial resolution of Stage 4: stride 32. After upsampling, all features have the same spatial shape, given as $[B, 2048, H, W]$, with $B$ the batch size and $H \times W$ the height and width at this resolution.

\subsubsection{Scale Attention Weighting}

Scale importance is indicated by per-scale attention weights that are learned. A mean feature map is computed by averaging the four projected features:

\begin{equation}
\bar{F} = \frac{1}{4} \sum_{i=1}^{4} F_i'
\end{equation}

A lightweight attention module processes $\bar{F}$, which consists of global average pooling followed by a small multilayer perceptron (MLP). More specifically, reducing $\bar{F}$ by global pooling passes through two fully connected layers with ReLU activation and a sigmoid activation at the output to give per-scale weights:

\begin{equation}
w = \text{Sigmoid}(\text{FC}_{512}(\text{ReLU}(\text{FC}_{2048}(\text{GAP}(\bar{F})))))
\end{equation}

This module outputs four scalar weights, corresponding to a ResNet stage and constrained to the range $[0, 1]$. These weights are independent and are not normalized to sum up to unity. \footnote{Implemented using $1 \times 1$ convolutions, which are functionally equivalent to fully connected layers after global average pooling.}

\subsubsection{Weighted Feature Summation}

The outputs of all four projections are summed in a weighted fashion to yield the fused feature map:

\begin{equation}
F_{\text{fused}} = \sum_{i=1}^{4} w_i \cdot F_i'
\end{equation}

The resulting map retains the shape $[B, 2048, H, W]$ and embodies an adaptive combination of information from all ResNet stages.

\subsection{Semantic Clustering with Rule-Based Pseudo-Labels}

Rule-based pseudo-labels are generated to provide coherent representations both spatially and semantically for every spatial location \cite{cho2022part}. This supplies a regularization signal without using teacher-student distillation methods.

\subsubsection{Spatial Partitioning}

Vertical partitioning is done based on anatomical priors for human bodies. Each spatial location with vertical coordinate $y \in [0, 1, 2]$ is assigned a semantic class:

\begin{equation}
\text{spatial\_class}(y) = \begin{cases}
0 & \text{if } y < 0.4 \text{ (upper body)} \\
1 & \text{if } 0.4 \leq y < 0.8 \text{ (lower body)} \\
2 & \text{if } y \geq 0.8 \text{ (shoes)}
\end{cases}
\end{equation}

This partition gives three regions of interest corresponding to the upper body (top 40\%), lower body (middle 40\%), and footwear (bottom 20\%).

\begin{figure}[h]
    \centering
    \includegraphics[width=1.0\linewidth]{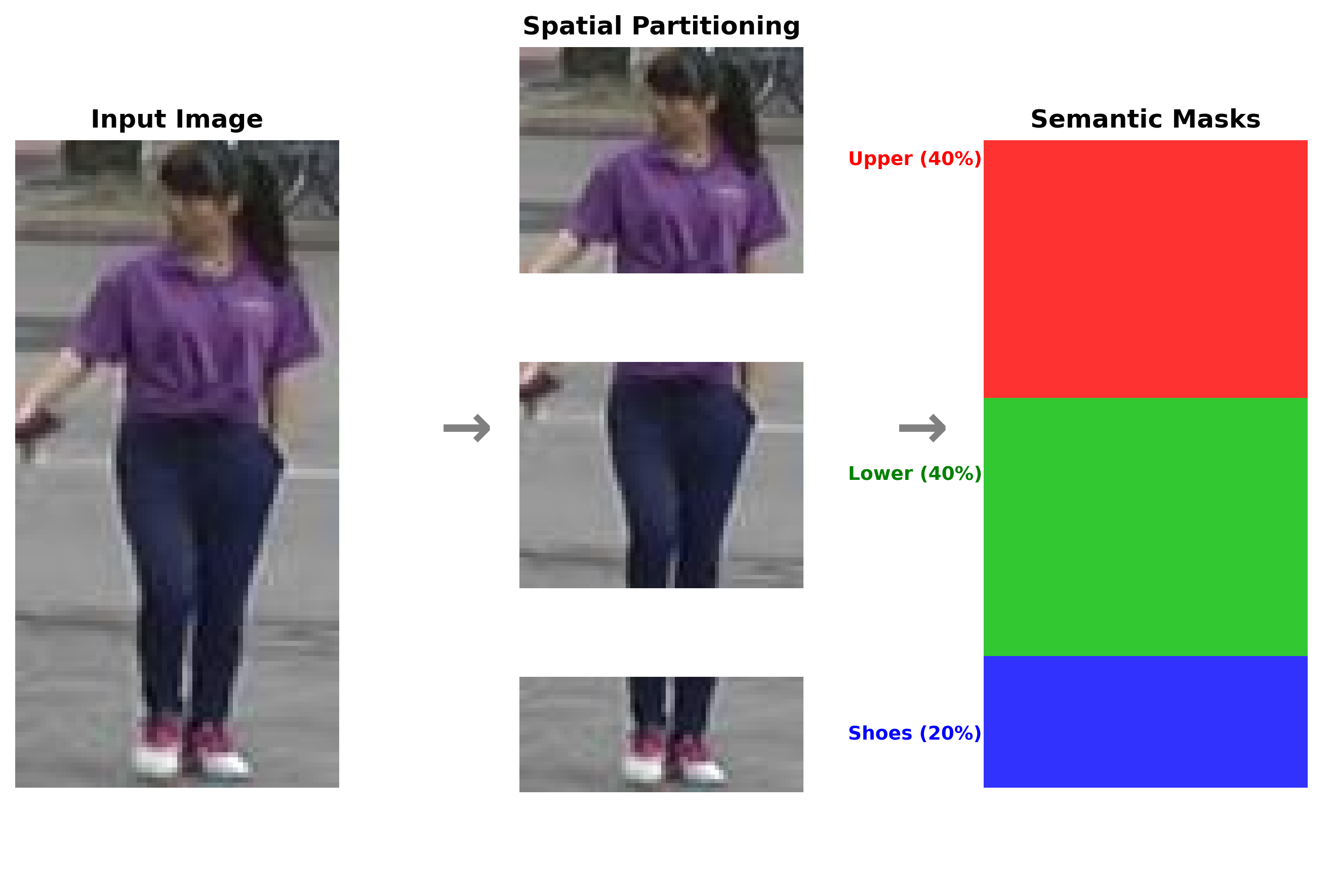}
    \caption{Spatial Clustering}
    \label{fig:single_col}
\end{figure}

\subsubsection{Foreground-Background Separation}

The foreground, or the person, is separated from the background by calculating the magnitude of the L2 norm of the fused feature at every spatial position. These magnitudes are then averaged, and their standard deviation is calculated across all positions:

\begin{equation}
\text{magnitude}(y, x) = \|F_{\text{fused}}(y, x)\|_2
\end{equation}

\begin{equation}
\mu = \text{mean}(\text{magnitude}), \quad \sigma = \text{std}(\text{magnitude})
\end{equation}

The location is classified as foreground if:

\begin{equation}
\text{magnitude}(y, x) > \mu + 0.5\sigma
\end{equation}

\subsubsection{Pseudo-Label Generation}

The final pseudo-label for each location contains the spatial class and the foreground-background information:

\begin{equation}
\text{pseudo\_label}(y, x) = \begin{cases}
\text{spatial\_class}(y) & \text{if foreground} \\
3 & \text{if background}
\end{cases}
\end{equation}

Thus, four semantic classes can be defined: upper body, lower body, shoes, and background.

\subsubsection{Semantic Classification Head}

The per-pixel semantic classification head takes in the fused feature map. The features are reshaped such that each spatial location represents a separate sample, and a compact neural network predicts the semantic class of each location.

We design the semantic head as a 3-layer multilayer perceptron: input size 2048, hidden layers with 1024 and 512 neurons, and an output layer with 4 neurons. Batch normalization and ReLU activations are used between layers, dropout is also applied at a rate of 0.1 for regularization. The semantic logits are given by:

\begin{equation}
\text{semantic\_logits} = \text{MLP}(F_{\text{fused\_flat}})
\end{equation}

The semantic classification loss is the cross-entropy between the predicted logits and pseudo-labels:

\begin{equation}
L_{\text{semantic}} = \text{CrossEntropy}(\text{semantic\_logits}, \text{pseudo\_labels})
\end{equation}

\subsection{Identity Classification Head}

Meanwhile, the fused feature map is used for identity classification. Global average pooling reduces the spatial dimensions to a $1 \times 1$ feature vector of size 2048:

\begin{equation}
\mathbf{f} = \text{GAP}(F_{\text{fused}})
\end{equation}

Next comes a batch-normalization layer, without a bias term, as is common in person re-identification:

\begin{equation}
\mathbf{f}_{\text{bn}} = \text{BN}(\mathbf{f})
\end{equation}

A linear classifier maps $\mathbf{f}_{\text{bn}}$ to the number of identities in the training set; for Market-1501 it is 751:

\begin{equation}
\text{logits}_{\text{id}} = \text{Linear}(\mathbf{f}_{\text{bn}})
\end{equation}

During inference, the batch-normalized features are used as embeddings for re-identification, while unit normalization and Euclidean distance are used to handle query-gallery matching.

\subsection{Loss Functions and Multi-Task Learning}

VisNet is trained with three complementary losses dynamically balanced. The losses are detailed next.

\subsubsection{Identity Classification Loss}

The identity classification branch uses the standard cross-entropy loss on the identity predictions:

\begin{equation}
L_{\text{CE}} = \text{CrossEntropy}(\text{logits}_{\text{id}}, y_{\text{id}})
\end{equation}

where $y_{\text{id}}$ contains the true identity labels.

\subsubsection{FIDI Metric Learning Loss}

FIDI provides a substitute for conventional triplet loss towards metric learning. FIDI frames metric learning as symmetric divergence minimization between a learned distribution $\mathcal{U}$ and a ground-truth-based distribution $\mathcal{K}$.

The FIDI loss is built upon relative entropy, a measure of the distance between two distributions. Let $\mathcal{K}$ be a known distribution of training image pairs, i.e., the ground truth identity labels, and $\mathcal{U}$ be an unknown distribution we aim to learn, then the FIDI loss is defined as follows~\cite{FIDI}:

\begin{equation}
L_{\text{FIDI}} = D(\mathcal{U}\|\mathcal{K}) + D(\mathcal{K}\|\mathcal{U})
\label{eq:fidi_symmetric}
\end{equation}

where the alpha-divergence is given by:

\begin{equation}
D(\mathcal{U}\|\mathcal{K}) = \sum_{p_{ij} \in \mathcal{P}} u_{p_{ij}} \log \frac{\alpha u_{p_{ij}}}{(\alpha - 1)u_{p_{ij}} + k_{p_{ij}}}
\label{eq:fidi_divergence}
\end{equation}

Here, $p_{ij} = \{\mathbf{x}_i, \mathbf{x}_j\}$ is a pair of image samples and $\mathcal{P}$ is a collection of image pairs. $k_{p_{ij}} \in \mathcal{K}$ and $k_{p_{ij}} = 1$ if the image pair $\mathbf{x}_i$ and $\mathbf{x}_j$ are from the same identity, and $k_{p_{ij}} = 0$ otherwise. $u_{p_{ij}}$ is taken from an unknown distribution $\mathcal{U}$, which is the distribution of feature level relationship of image pairs in $\mathcal{P}$.

\subsubsection{Semantic Clustering Loss}

The semantic clustering loss corresponds to the cross-entropy loss on the per-pixel semantic classification task:

\begin{equation}
L_{\text{semantic}} = \text{CrossEntropy}(\text{semantic\_logits}, \text{pseudo\_labels})
\end{equation}

\subsubsection{Total Loss with Dynamic Weight Averaging}

Instead of fixed weights, the three losses balance each other using dynamic weight averaging (DWA) \cite{LiuEtAl2019}:

\begin{equation}
L_{\text{total}} = w_{\text{FIDI}}(t) \cdot L_{\text{FIDI}} + w_{\text{CE}}(t) \cdot L_{\text{CE}} + w_{\text{semantic}}(t) \cdot L_{\text{semantic}}
\end{equation}

where weights $w_i(t)$ are computed at batch $t$ based on recent loss histories.

\subsection{Training Strategy: Dynamic Weight Averaging and Batch Sampling}
\label{sec:dwa}

DWA adjusts weights based on the rate of decrease for each loss. For each task, it keeps track of the loss values over the last 50 batches. The ratio $r_i(t) = \frac{L_i(t)}{L_i(t-1) + \epsilon}$ is computed at batch $t$. The weights are computed using a softmax normalization with temperature $T = 2.0$:

\begin{equation}
w_i(t) = \frac{\exp(r_i(t) / T)}{\sum_j \exp(r_j(t) / T)}
\end{equation}

This automatically increases the relative influence of tasks that are currently improving more slowly, avoiding domination by the most rapidly improving objective. Temperature determines the smoothness of weight changes across batches.

It also employs PK sampling\cite{liao2022graph,hermans2017defence,schroff2015facenet}, with each batch having $P$ identities and $K$ images per identity. More specifically, we use $P = 8$, $K = 12$, and a final batch size of 96 samples; this can also help ensure diversity within batches and support hard negative mining. All three losses are turned on starting from epoch 0, which corresponds to single-stage training. The DWA mechanism self-balances the objectives without requiring manual tuning. During testing, L2-normalized features are extracted from the identity classification head. Re-identification is treated as a retrieval task, computing distances between query and gallery images, followed by ranking by similarity. We report two standard metrics, Cumulative Matching Characteristic (CMC) and Mean Average Precision mAP.

\subsection{Semantic-Aware Data Augmentation Pipeline}
\label{subsec:augmentation}

To improve model robustness against background variation, we implement a semantic segmentation-based augmentation pipeline. This module decouples the foreground subject from the background, allowing for targeted background perturbations while preserving the person's identity features.

The pipeline utilizes {YOLOv8-Seg} for real-time instance segmentation to generate binary masks $M \in \{0, 1\}^{H \times W}$, isolating the person pixels $I_{person} = I \odot M$. We then apply a stochastic transformation function $\mathcal{T}(\cdot)$ exclusively to the background region $I_{bg} = I \odot (1 - M)$. The final augmented image $I_{aug}$ is reconstructed as:

\begin{equation}
    I_{aug} = I_{person} + \mathcal{T}(I_{bg})
\end{equation}

We perform six distinct transformation categories to synthesize diverse environmental conditions:

\begin{enumerate}
    \item Color Space Manipulation: Operating in the HSV domain, we apply random hue rotation ($\theta \in [30^{\circ}, 150^{\circ}]$), saturation scaling ($\alpha \in [1.0, 2.0]$), and brightness modulation ($\beta \in [0.7, 1.3]$) to simulate varying lighting conditions.
    
    \item Texture Synthesis: To reduce reliance on background texture cues, we apply edge enhancement via Canny edge detection and emboss filtering kernels.
    
    \item {Noise Injection:} We introduce stochastic noise, including salt-and-pepper noise and additive Gaussian noise ($\mu=0, \sigma^2 \in [0.01, 0.05]$), to simulate sensor degradation.
    
    \item {Blur Simulation:} Motion blur and radial zoom blur kernels are convolved with the background to mimic camera motion and depth-of-field effects.
    
    \item {Pattern Generation:} Synthetic geometric structures, including grid lines, concentric circles, and diagonal stripes, are rendered with randomized colors to force the model to ignore structured background clutter.
    
    \item {Gradient Fill:} We generate linear, radial, and angular gradients with random start/end colors to simulate smooth, non-textured environments.
\end{enumerate}

Each augmentation technique is applied with a configurable probability $p$ and intensity strength $\lambda \in [0.0, 1.0]$, ensuring diverse training samples that reinforce the model's focus on person-specific features rather than environmental context.

\section{Experiments}
\label{sec:experiments}

\subsection{Dataset and Evaluation Metrics}

Experiments are conducted on Market-1501, containing 1,501 identities, 42,872 (originally 32,668) augmented training images, 3,369 query images, and 19,733 gallery images. The input images are resized to $256 \times 128$, padded by 10 pixels on all sides, then random crop back to $256 \times 128$. It also includes random horizontal flip with probability 0.5 and color jitter with brightness 0.2, contrast 0.15, saturation 0.15, and hue 0.1. Moreover, random erasing with probability 0.5 that affects 2--40\% of image area and normalized using ImageNet statistics. Only resizing and normalization are done at the time of testing. Following standard protocol, we exclude same-camera matches. We report Cumulative Matching Characteristic (CMC) at Rank-1, Rank-5, Rank-10, Rank-20, and mean Average Precision (mAP), computed without re-ranking, as shown in Table~\ref{tab:accuracy}.

\begin{table}[t]
\centering
\caption{Accuracy Performance of VisNet on Market-1501 trained from scratch.}
\label{tab:accuracy}
\begin{tabular}{l|c|cccccc}
\hline
\textbf{Method} & \textbf{Year} & \textbf{R1} & \textbf{R5} & \textbf{R10} & \textbf{R20} & \textbf{mAP} \\
& & \textbf{(\%)} & \textbf{(\%)} & \textbf{(\%)} & \textbf{(\%)} & \textbf{(\%)} \\
\hline
TransReID & 2021 & 95.20 & 98.0 & 98.7 & 99.1 & 89.50 \\
CLIP-ReID & 2023 & 88.10 & 93.8 & 95.6 & 96.9 & 80.30 \\
\textbf{VisNet (Ours)} & \textbf{2025} & \textbf{87.05} & \textbf{93.18} & \textbf{95.90} & \textbf{97.15} & \textbf{77.65} \\
AANet & 2019 & 82.60 & 90.5 & 93.2 & 95.1 & 72.20 \\
IDE\cite{zheng2018discriminatively} & 2018 & 79.51 & - & - & - & 59.87 \\
PSE\cite{sarfraz2018pose} & 2018 & 87.7 & 94.5 & 96.8 & - & 69.0 \\
TriNet\cite{hermans2017defense} & 2017 & 84.9 & - & - & - & 69.1 \\
SVDNet\cite{sun2017svdnet} & 2017 & 82.3 & - & - & - & 62.1 \\
MGN(flip)\cite{wang2018learning} & 2018 & 95.7 & - & - & - & 86.9\\
\hline
\end{tabular}
\end{table}

While TransReID achieves 95.20 percent Rank-1 accuracy, it relies on Vision Transformer pre-training. Similarly, CLIP-ReID leverages large-scale vision-language pre-training from CLIP. VisNet achieves 87.05 percent Rank-1 accuracy when trained from scratch on Market-1501 only, demonstrating strong performance without external pre-training. Our method outperforms AANet, a method that also uses a standard ResNet50 backbone, validating the effectiveness of multi-scale fusion and semantic clustering.

\begin{table}[t]
\centering
\caption{Computational Efficiency Analysis. VisNet achieves the best efficiency trade-off among methods with competitive accuracy (82.6\%+ Rank-1): 18.91 accuracy points per GFLOP, significantly outperforming TransReID (5.36) and CLIP-ReID (7.34).}
\label{tab:efficiency}
\begin{tabular}{l|cccc}
\hline
\textbf{Method} & \textbf{Params (M)} & \textbf{FLOPs (G)} & \textbf{R1 (\%)} & \textbf{Acc/GFLOP} \\
\hline
\textbf{VisNet (Ours)} & \textbf{32.41} & \textbf{4.601} & \textbf{87.05} & \textbf{18.91} \\
CLIP-ReID & 63.0 & 12.0 & 88.10 & 7.34 \\
TransReID & 86.0 & 17.8 & 95.2 & 5.36 \\
AANet & 19.0 & 2.5 & 82.60 & 33.04 \\
MGN (flip) & 68.75 & 48 & 95.7 & 1.99 \\
\hline
\end{tabular}
\end{table}

\subsection{Ranking Quality Across Metrics}

Beyond Rank-1, VisNet demonstrates strong performance across ranking metrics: 93.18\% (Rank-5), 95.90\% (Rank-10), 97.15\% (Rank-20). This indicates well-calibrated rankings are valuable for practical deployment where users review multiple candidates.

\begin{table}[t]
\centering
\caption{Model Component Analysis: VisNet Parameter Distribution of VisNet's Model}
\label{tab:components}
\begin{tabular}{l|c|c}
\hline
\textbf{Model Component} & \textbf{Parameters} & \textbf{Percentage} \\
\hline
ResNet50 Backbone (Stage 0-4) & 23,508,032 & 72.57\% \\
Multi-scale Fusion & 4,733,444 & 14.61\% \\
Semantic Clustering Head & 2,628,100 & 8.11\% \\
Classifier & 1,538,048 & 4.75\% \\
BN Neck & 4,096 & 0.01\% \\
\hline
\textbf{Total} & \textbf{32,411,720} & \textbf{100.00\%} \\
\hline
\end{tabular}
\end{table}

The parameter breakdown in Table~\ref{tab:components} reveals VisNet's design philosophy. The ResNet50 backbone comprises 72.57\% of total parameters, while our proposed multi-scale fusion (14.61\%) and semantic clustering (8.11\%) components add efficient regularization mechanisms. The identity classifier contributes 4.75\% of parameters for the final person re-identification task. This architecture demonstrates that competitive accuracy can be achieved through intelligent module design rather than replacing the backbone with lightweight alternatives, resulting in a total model size of 32.41M parameters.

Experiments confirm that VisNet is competitively accurate, achieving a Rank-1 score of 87.05\% using a standard backbone, which is 4.45\% higher than AANet. More importantly, Table~\ref{tab:efficiency} shows that VisNet is efficient, achieving 18.91 accuracy points per GFLOP, over $3.5\times$ more efficient compared to TransReID at 5.36. While AANet scores 33.04, it only achieves 82.60\% R1. The streamlined design, where semantic modules add only 22.7\% parameter overhead to the ResNet50 base, makes VisNet amenable to resource-constrained deployment without performance degradation in ranking.

\section{Conclusion}
\label{sec:conclusion}

VisNet achieves competitive accuracy while maintaining efficiency-accuracy trade-off through multi-scale fusion of learned attention into each individual scale's feature representation, coupled with guiding clustering with pseudo-labels. Our efficient method demonstrates that it is possible to reach high accuracy with an efficient and lightweight approach while reducing 3.87$\times$ GFLOPs as compared to the standard benchamrk models and can therefore be used where other approaches are too heavy or when there is limited availability of resources.

\section{Future Work}
\label{sec:future}
A combination of learned semantic prototypes with teacher–student distillation may improve semantic understanding of body divisions. Moreover, shifting from scale attention to self-attention may strengthen connections among body regions.

\bibliographystyle{IEEEtran}
\bibliography{references}

\end{document}